\documentclass[%
 aip,
 amsmath,amssymb,
preprint,
]{revtex4-1}

\usepackage{graphicx}
\usepackage{dcolumn}
\usepackage{bm}

\usepackage[utf8]{inputenc}
\usepackage[T1]{fontenc}
\usepackage{mathptmx}
\usepackage{etoolbox}
\usepackage{algorithmic}
\RequirePackage{algorithm}[0.1]

\makeatletter
\def\@email#1#2{%
 \endgroup
 \patchcmd{\titleblock@produce}
  {\frontmatter@RRAPformat}
  {\frontmatter@RRAPformat{\produce@RRAP{*#1\href{mailto:#2}{#2}}}\frontmatter@RRAPformat}
  {}{}
}%
\makeatother

\begin{document}

\preprint{AIP/123-QED}

\title[Score-Based Model Assisted Sampling]{Micro-Macro Consistency in Multiscale Modeling: \\ Score-Based Model Assisted Sampling of Fast/Slow Dynamical Systems}
\author{E. R. Crabtree}
\author{J. M. Bello-Rivas}%
 \author{I. G. Kevrekidis}
\affiliation{ 
Department of Chemical and Biomolecular Engineering, Johns Hopkins University, Baltimore, MD 21218 USA 
}%

\date{\today}

\begin{abstract}
A valuable step in the modeling of multiscale dynamical systems in fields such as computational chemistry, biology, materials science and more, is the representative sampling of the phase space over long timescales of interest; this task is not, however, without challenges. For example, the long term behavior of a system with many degrees of freedom often cannot be efficiently computationally explored by direct dynamical simulation; such systems can often become trapped in local free energy minima. In the study of physics-based multi-time-scale dynamical systems, techniques have been developed for enhancing sampling in order to accelerate exploration beyond free energy barriers.
On the other hand, in the field of Machine Learning, a generic goal of generative models is to sample from a target density, after training on empirical samples from this density. Score based generative models (SGMs) have demonstrated state-of-the-art capabilities in generating plausible data from target training distributions. Conditional implementations of such generative models have been shown to exhibit significant parallels with long-established --and physics based-- solutions to enhanced sampling. These physics-based methods can then be enhanced through coupling with the ML generative models, complementing the strengths and mitigating the weaknesses of each technique. In this work, we show that that SGMs can be used in such a coupling framework to improve sampling in multiscale dynamical systems.
\end{abstract}

\maketitle

\begin{quotation}
We discuss the parallels between generative models in machine learning, particularly score-based generative models (SGMs), and traditional physics-based enhanced sampling methods like Umbrella Sampling (US). We show that SGMs can be coupled with traditional enhanced sampling to mitigate drawbacks that each tool has when used alone. We apply this coupling method to multi-time-scale stochastic dynamical systems and demonstrate that the coupling can offer faster convergence to the desired system conditional distribution.
\end{quotation}

\section{Introduction}
\label{sec:intro}
In the realm of computational chemistry and physics, numerous enhanced sampling tools have been developed for the purpose of exploring the phase space of multiscale systems of interest \cite{HeninLelievreShirtsValssonDelemotte2022}. Often, the goal in the implementation of these tools is to understand system behavior over long time scales; this is challenging due to the many degrees of freedom in typical models and simulations that are often performed at the fine scale. Many of these enhanced sampling tools are aimed at coarse-grained, low-dimensional models, in terms of coarse collective variables, that describe the effective, macroscopic behavior of the system. When these reduced descriptors are not \emph{a priori} known, they must be identified, either analytically or in a data-driven way, before enhanced sampling can proceed. The construction of reduced, coarse-scale models from fine scale ones, in terms of such ``Collective Variables'' (CVs), has been a prevalent question in multiscale modeling for decades, and many data-driven techniques for identifying low dimensional descriptors have been developed over the years \cite{pca, autoencoder, isomap, locallinear, COIFMANdmaps}. It follows that, once a reduced order representation of a system is constructed, there also arises a need for ``returning'' selected reduced representation instances back to corresponding instances in the full, ambient space of the system. The task is not trivial, but methods to practically accomplish this inverse mapping, sometimes called ``lifting'' \cite{cms_eq_free,eq_free_algs_apps} back to the ambient space, have been developed and commonly used in many fields involving first principles modeling. For example, umbrella sampling (US) in molecular dynamics \cite{TORRIE1977} involves adding a harmonic biasing potential to a CV in order to bias a fine-scale molecular simulation toward prescribed coarse-scale variable values of interest. This produces fine-scale data {\em conditioned on a coarse-scale representation}, and thus accomplishes the aforementioned task of ``lifting.''

Meanwhile, in the field of Machine Learning, generative models have become popular for their ability to generate high dimensional data from latent distributions \cite{goodfellow2014, VAE, mirza2014}. In recent years, generative adversarial networks (GANs) have been used to generate realistic data (such as images of a human face) from partial descriptors (e.g. eyes, noses); furthermore, both GANs and other neural network based generative models have been used for the purpose of ``lifting'' in the context of returning lower dimensional, coarse-grained descriptions back to the full ambient space of a multiscale dynamical system --whether through the interpolation of molecular conformations, or through generation of high dimensional data from identified CV representations \cite{mlss, stoltz2022}. Presently, score based generative models (SGMs) constitute state-of-the-art models that have demonstrable capabilities in generating plausible data samples from a target training distribution in complex problems and systems \cite{songdiffmodels, diffmodels2}.

Established enhanced sampling methods, whose outputs are fine-scale representations consistent with (conditioned on) coarse-scale representations, present parallels with new and emerging ML generative models. Both categories of techniques accomplish similar tasks, but exhibit contrasting drawbacks: physics based sampling methods (that do not require training data) often suffer from slow convergence due to metastability; generative models must be trained offline on huge datasets and, by virtue of being data-driven, they will not explicitly account for all physical constraints of a system. We will demonstrate that the two types of methods, arising from differing origins, are capable of complementing each other, potentially alleviating their respective drawbacks, for more efficient sampling when used in tandem. Previous work demonstrated that {\em conditional GANs} can be used efficiently to create new microscopic initial conditions for physics-based enhanced sampling tools; conversely, the enhanced ML sampling can create new training data for the GANs on-the-fly \cite{GANs_closures}. The present work shows that this coupling framework can be extended to new state-of-the-art generative models, SGMs.

The paper is organized as follows: Section \ref{sec:overview} presents an overview of SGMs as a generative model. Section \ref{sec:alg} proposes a framework in algorithmic form for coupling SGMs with physics-based enhanced sampling methods. Section \ref{sec:SGM_approx_comp} demonstrates the capabilites of SGMs in generating conditional distributions from training data produced by multiscale stochastic dynamical systems models. Section \ref{sec:SGM_ICs} contains illustrative results from applying the coupling framework to different examples. Section \ref{sec:conclusions} critically summarizes the results and outlines possible future work.

\section{Overview of SGMs}
\label{sec:overview}
\subsection{Denoising SDEs}
SGMs are trained to generate new data points of a target density by employing a discretized stochastic differential equation (SDE): this SDE takes the form of a reverse, or backward, SDE of a forward SDE of choice. The forward SDE is any SDE that 
would gradually transform training data into normally distributed noise, while the reverse SDE gradually ``denoises'' a normally distributed variable back to plausible data. These SDEs can take many possible forms but, in generic form, can be defined as

\begin{equation}
    \mbox{d} x = f(x,t) \mbox{d} t + g(t)\mbox{d} B_t,
\end{equation}

\noindent
and the corresponding reverse SDE is defined as

\begin{equation}
    \mbox{d} x = [f(x,t) - g(t)^2\nabla_x \log p_t(x)]\mbox{d} t + g(t)\mbox{d} B_t.\\
    \label{eq:reversesde}
\end{equation}

A feature of these two SDEs is that, if the initial condition in the reverse SDE is chosen to be $x(T)$ of the forward SDE (where $T$ is some specified time) then the reverse SDE can be integrated to eventually obtain $x(0)$. Therefore, if we are able to run the reverse SDE initialized from the distribution $x(T)$, then we can generate samples from $x(0)$. We do not know $x(0)$ \emph{a priori}; however, we do have access to $x(T)$ through the feature of choosing or parameterizing the SDE, so that its marginals with respect to time (here these marginals of $x$ with respect to time, $t$, are written as $p_t(x)$, and when this marginal refers to a specific time, for example $t = 0$, this marginal can be written as $p(x(0))$) converge to a normal distribution at an exponential speed, independently of the original distribution of $x(0)$. Therefore, we can start the reverse SDE in $x(T) = \mathcal{N}(0, I)$. The last unknown of this problem is then $x(t)$ and therefore the drift $\nabla \log p_t(x)$ (where again $p_t(x)$ is the marginal distribution of $x$ at time $t$) of the reverse SDE, which we can then approach by training a neural network (whose inputs are the samples $x$ and respective times $t$, with learnable parameters denoted by $\theta$) to approximate the unknown $s_\theta(x,t) \approx \nabla \log p_t(x)$.

\subsection{Neural Network Drift Approximation}
The reverse SDE is implemented using the Euler-Maruyama method \cite{maruyama1955continuous}. To advance the SDE by $\delta t = t_{i+1} - t_i$, we use the following discretized equation:
\begin{equation}
    x(t_{i+1}) = x(t_i) + (t_{i+1} - t_i) \left( f(x,t_i) - g(t)^2\nabla_x \log p_{t_i}(x)\right) + \sqrt{(t_{i+1} -  t_i)} g(t)B_{t_{i+1} - t_i},
\end{equation}
where $B_{t_{i+1} - t_i}$ is a normal random variable with standard variance.
Now that we have a discretized reverse SDE, we can use a neural network to approximate $\nabla_x \log \hat{p}_t(x)$ (where $\hat{p}_t(x)$ refers to an approximation of the distribution ${p}_t(x)$). We do this through Score-Matching \cite{songdiffmodels, songscore, vincent2011connection}, resulting in the following loss for the neural network approximation of the reverse SDE drift:
$${L}(\theta) = \mathbb{E}_{t \sim U(0,T)}\mathbb{E}_{x(0) \sim p(x(0))}\mathbb{E}_{x(t) \sim p(x(t) | x(0))}[\lambda (t) \| \nabla \log \hat{p}_t(x) - s_\theta(x(t), t)\|^2_2],$$
where $U(0,T)$ is a uniform distribution over the time interval of [0,T], $\lambda (t)$ is a positive weighting function, and $x(0)$ denotes $x(t)$ at time $t=0$. In practice this Score-Matching loss is comparable to a more computationally efficient \emph{denoising score matching objective} \cite{vincent2011connection}, which can be used as a substitute loss function.
\subsection{Conditional Diffusion Models}
We can also create a conditional diffusion model by having the neural network approximate $\nabla \log \hat{p}_t(x|y)$ where $y$ is some given label, using the following score-matching loss function, also called the conditional denoising estimator (CDE), used and discussed in previous works \cite{tashiro2021csdi,saharia2021image, batzolis2021conditional}:
$${L}(\theta) = \mathbb{E}_{t \sim U(0,T)} \mathbb{E}_{x(0),y \sim p(x(0) | y)} \mathbb{E}_{x(t) \sim p(x(t) | x(0))}[\lambda (t)\| \nabla \log \hat{p}_t(x | y) - s_\theta(x(t), t, y)\|^2_2],$$
meaning that the new reverse SDE that we will evaluate becomes:
\begin{equation}
    x({t_{i+1}}) = x({t_i}) + (t_{i+1} - t_i) \left( f(x,t) - g(t)^2\nabla_x \log p_t(x|y)\right) + \sqrt{t_{i+1} - t_i} g(t)B_{t_{i+1} - t_i}.
\end{equation}
This is essentially the same SDE as equation 3 (the unconditional reverse SDE), but with a change to the Neural Net learned drift term.
For conditional SGMs, Song et al. \cite{songdiffmodels} originally proposed a SDE with a score matching term of
\begin{equation}
    \nabla_x \log p_t(x|y) = \nabla_x \log p_t(x) + \nabla_x \log p_t(y|x).
    \label{eq:song_cond_term}
\end{equation}
While this term may be convenient \emph{for situations in which an unconditional SGM has already been trained on the system of interest (or if the heuristics and domain knowledge needed to construct the closed form of $\log p_t(y|x)$ are readily available)}, we are only interested in training conditional SGMs (cSGMs) in this work. Therefore, we choose to use a reverse SDE with a score matching term that only requires training to approximate $\nabla_x \log p_t(x|y)$, and does not require prior training of an unconditional model in addition to an approximation of $\nabla_x \log p_t(y|x)$. Note that in this work, we are conditioning on continuous variables, rather than discrete labels. Diffusion models in literature have typically used discrete labels for class conditional generation \cite{songdiffmodels}, while continuous labels have been used for regression based generation purposes such as inpainting \cite{songdiffmodels, tashiro2021csdi, saharia2021image, batzolis2021conditional}. 

\section{Proposed Framework}
\label{sec:alg}
Algorithms 1 and 2 lay out the general framework of our proposed method for coupling SGMs with physics-based sampling methods.
\begin{algorithm}[H]
\caption{Training a Conditional SGM on a Dynamical System}
\hspace*{\algorithmicindent} \textbf{Input:} Samples of data from a dynamical system, $x(t)$, with a corresponding distribution $P(x)$ and corresponding labels, $y$, if known \emph{a priori}
\begin{algorithmic}[1]
\STATE{Sample from $P(x)$};
\STATE{Identify a slow direction of the system, $y$,  through \emph{a priori} knowledge or using nonlinear dimensionality reduction};
\STATE{Train a cSGM on data sampled from $P(x)$ with labels, $y$, that correspond to the slow direction of the system};
\STATE{Use the trained cSGM to sample from the approximated conditional measures of the system, $\hat{P}(x|y)$}\\
\RETURN $\hat{P}(x|y)$ at prescribed $y$ values.
\label{alg:Training_SGM_alg}
\end{algorithmic}
\end{algorithm}

\begin{algorithm}[H]
\caption{Sampling Initialized by Conditional SGMs}
\hspace*{\algorithmicindent} \textbf{Input:} $N$ samples produced by a SGM from an estimated distribution, $\hat{P}(x|y)$ at prescribed $y$ values
\begin{algorithmic}[1]
\FOR{$n = 1$ \textbf{to} $N$} 
\STATE{Use physics-based biasing methods such as Umbrella Sampling to sample from $P(x|y)$ initialized by an initial condition generated from $\hat{P}(x|y)$ --for new values of $y$--}; 
\ENDFOR
\STATE Reweigh the biased samples if necessary
\RETURN $P(x|y)$ at interpolated and extrapolated $y$ values.
\label{alg:coupled_samp_alg}
\end{algorithmic}
\end{algorithm}

 While training the cSGM has an up-front computational cost, once trained it can provide samples from a reasonable estimation of the true conditional distribution at the prescribed values set by the user. These estimated samples can then be used to initialize methods like US and help them converge to the true conditional distribution of the system faster than the original method, without the ``informed'' initialization. Section \ref{sec:SGM_ICs} further elaborates with numerical examples and presents evidence that this proposed approach can be faster and more computationally efficient than biased sampling of the dynamical system alone (provided that the conditional approximations generated by the SGM are sufficient approximations of the true conditionals). Furthermore, all SDE parameters for our SGM models were chosen using guidance from Karras et. al. \cite{karrasdiffparams}. Network hyperparameters were also considered. The examples investigated in this work are fundamentally different from the image examples used by Karras, so our network is smaller and simpler than their networks. For all 2D dynamical system examples, we used a dense network of 8 layers of 64 to 512 nodes. Larger networks were also investigated but only provided marginal improvements.  

\section{Using SGMs to approximate conditional equilibrium measures}
\label{sec:SGM_approx_comp}
In the field of molecular simulation, many techniques have been created and used to surmount free energy barriers that impede the exploration of a molecule's phase space; full, efficient exploration of these phase spaces is important to fields such as drug design, materials discovery, and more. A popular example of this in computational chemistry is the aforementioned US method \cite{TORRIE1977}. In US, an artificial bias is added to the system to force the dynamics toward a prescribed value of a variable that governs the bulk behavior of the system, often called a collective variable (CV). Methods using machine learning-based generative models to produce high-dimensional molecular configurations consistent with given values for a system's CVs have been developed, and we have shown in previous work that coupling of the physics-based simulations with machine learning models is possible, using Generative Adversarial Networks as the ML model \cite{GANs_closures}. In this work, we propose a framework for coupling SGMs with physics-based simulations, initializing the physics with SGM generated data consistent with a desired CV.

However, before discussing the proposed method of coupling SGMs with physics based sampling, we first must discuss how well SGMs by themselves approximate conditional measures of multi-time-scale dynamical systems.
Consider a dynamical system described by a two-dimensional set of multiscale SDEs:
\begin{equation}
    \begin{split}
        \mbox{d} z_1(t) &= a_1\mbox{d} t + a_2\mbox{d} B_1 \\
        \mbox{d} z_2(t) &= -(-1 + 0.2z_1(t) + 4z_2(t)(-1 + z_2(t)^2))\mbox{d}t + a_3 \mbox{d} B_2(t)
    \end{split}
\end{equation}
where $B_1$ and $B_2$ are standard normal random variables, $a_1 = a_2 = 10^{-4}$, and $a_3 = \epsilon{a_1} = \epsilon{a_2} = 10^{-1}$ where the ratio of time scales is $\epsilon = 10^3$; this implies that $z_1$ is the slow variable and $z_2$ is the fast variable ($z_2$ is faster than $z_1$ by a factor of $10^3$). In addition to being the ``fast'' fine-scale variable in this toy example, the behavior of $z_2$ is defined by a double-well potential where two basins of attraction at $z_2 = 1$ and $z_2 = -1$ are separated by a tall barrier. Furthermore, the wells are slowly evolving in relative depth as the values of the slow direction, $z_1$, evolve in time. At $z_1 = 0$, the well at $z_2 = 1$ is relatively much deeper than the other, and as $z_1$ increases with time, the well at $z_2 = 1$ becomes progressively shallower, while the well at $z_2 = -1$ becomes deeper. The wells have the same depth when $z_1 = 5$, and the evolving well behavior is symmetric around $z_1 = 5$.
Analogous to US in molecular dynamics, performed by adding a harmonic bias to a collective variable, we can sample conditional distributions of the $z_2$ variable by adding a harmonic biasing potential to the $z_1$ equation of our system, resulting in the following new equations:
\begin{equation}
    \begin{split}
        \mbox{d} z_1(t) &= (a_1 - \kappa(z_1(t) - z^0_1))\mbox{d} t + a_2\mbox{d} B_1 \\
        \mbox{d} z_2(t) &= -(-1 + 0.2z_1(t) + 4z_2(t)(-1 + z_2(t)^2))\mbox{d}t + a_3 \mbox{d} B_2(t)
    \end{split}
\end{equation}
where $\kappa$ is a force constant dictating the strength of the bias, and $z^0_1$ is the value of $z_1$ toward which we wants to bias the simulation. We trained a cSGM on a long unbiased simulation of this system where the samples are labeled by their $z_1$ values (in this case, the dataset consisted of 100,000 samples from a simulation of these SDEs with $\mbox{d} t = 10^{-2}$). We then compared an output of the trained SGM with a US output conditioned at the same slow variable ($z_1$) value, as well as the true probability density function (PDF) given by the equations of the system.

\begin{figure}[ht]
    \centering
    \includegraphics[width=1\linewidth]{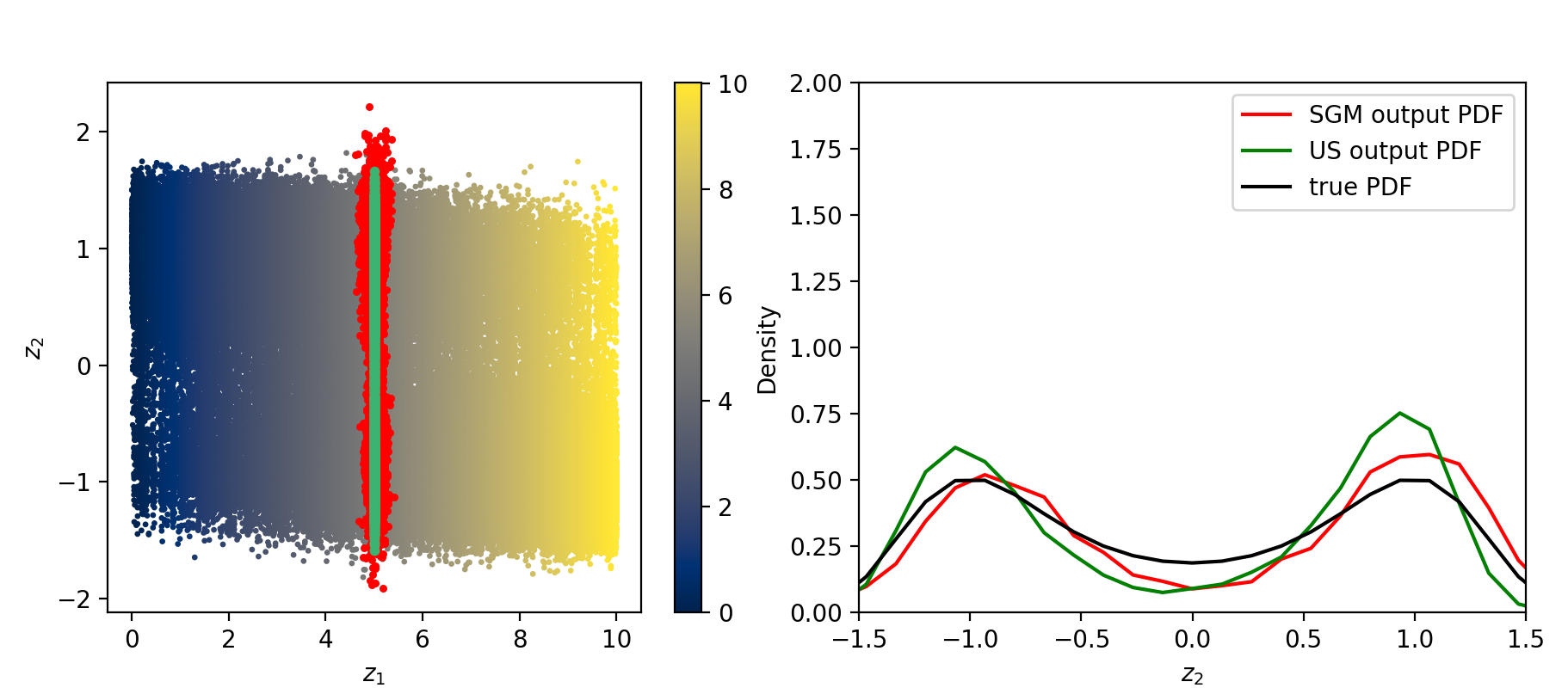}
    \caption{On the left, output from a trained SGM conditioned at $z_1 = 5$ (red) is compared with the output from a US simulation (green) conditioned on the same $z_1$ value on top of a background of the training data for the SGM. On the right, the PDF estimated from the SGM output is compared to the PDF estimated from the US simulation, as well as with the true PDF, given by the known $z_2$ equation of the system (black line).}
    \label{fig:SGM_and_US_comp_5}
\end{figure}

\begin{figure}[ht]
    \centering
    \includegraphics[width=0.9\linewidth]{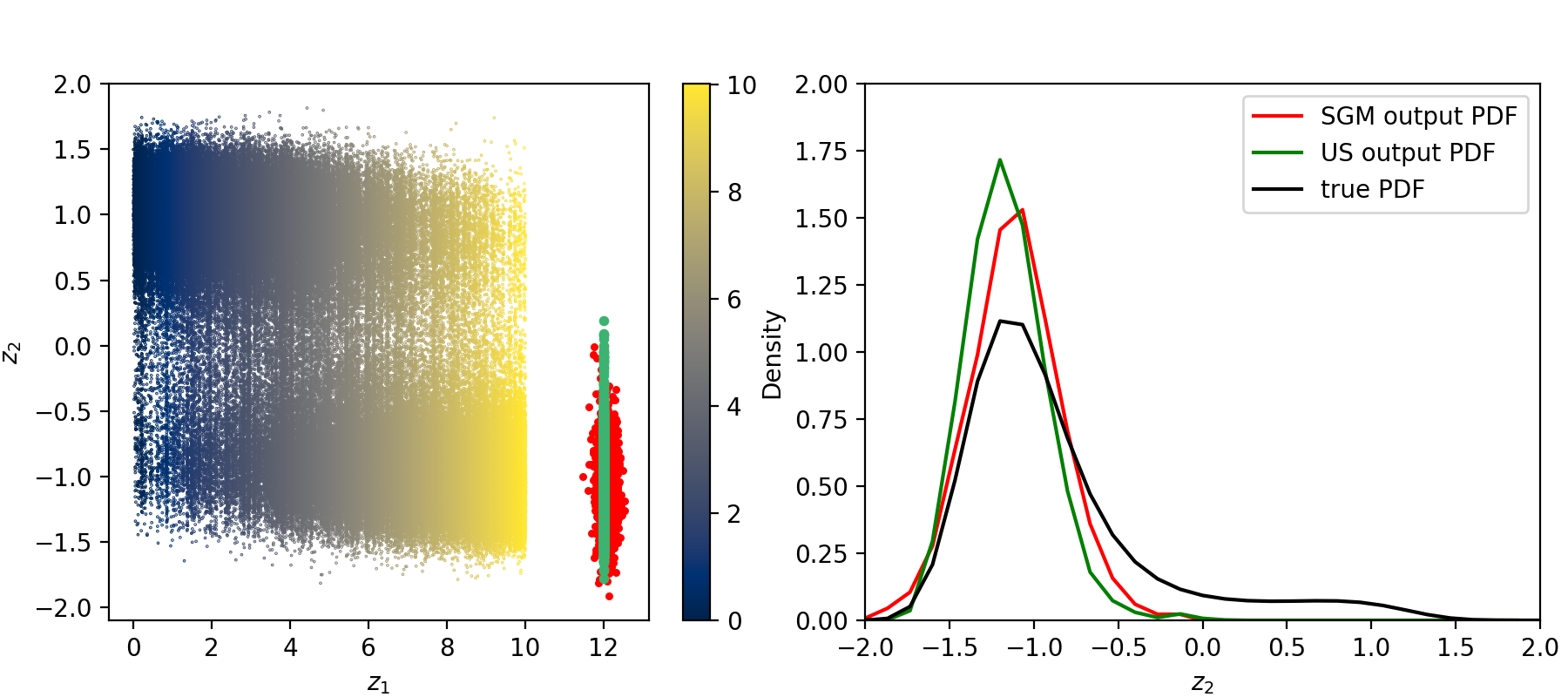}
    \caption{On the left: output from a trained SGM, conditioned at $z_1 = 12$ (red) is compared with the output from a US simulation (green), conditioned at the same $z_1$ value plotted with the training data for the SGM. On the right, the PDF estimated from the SGM output is compared to the PDF estimated from the US simulation, as well as with the true PDF given by the known $z_2$ equation of the system (black line).}
    \label{fig:SGM_and_US_comp_12}
\end{figure}

\begin{figure}[H]
    \centering
    \includegraphics[width=0.75\linewidth]{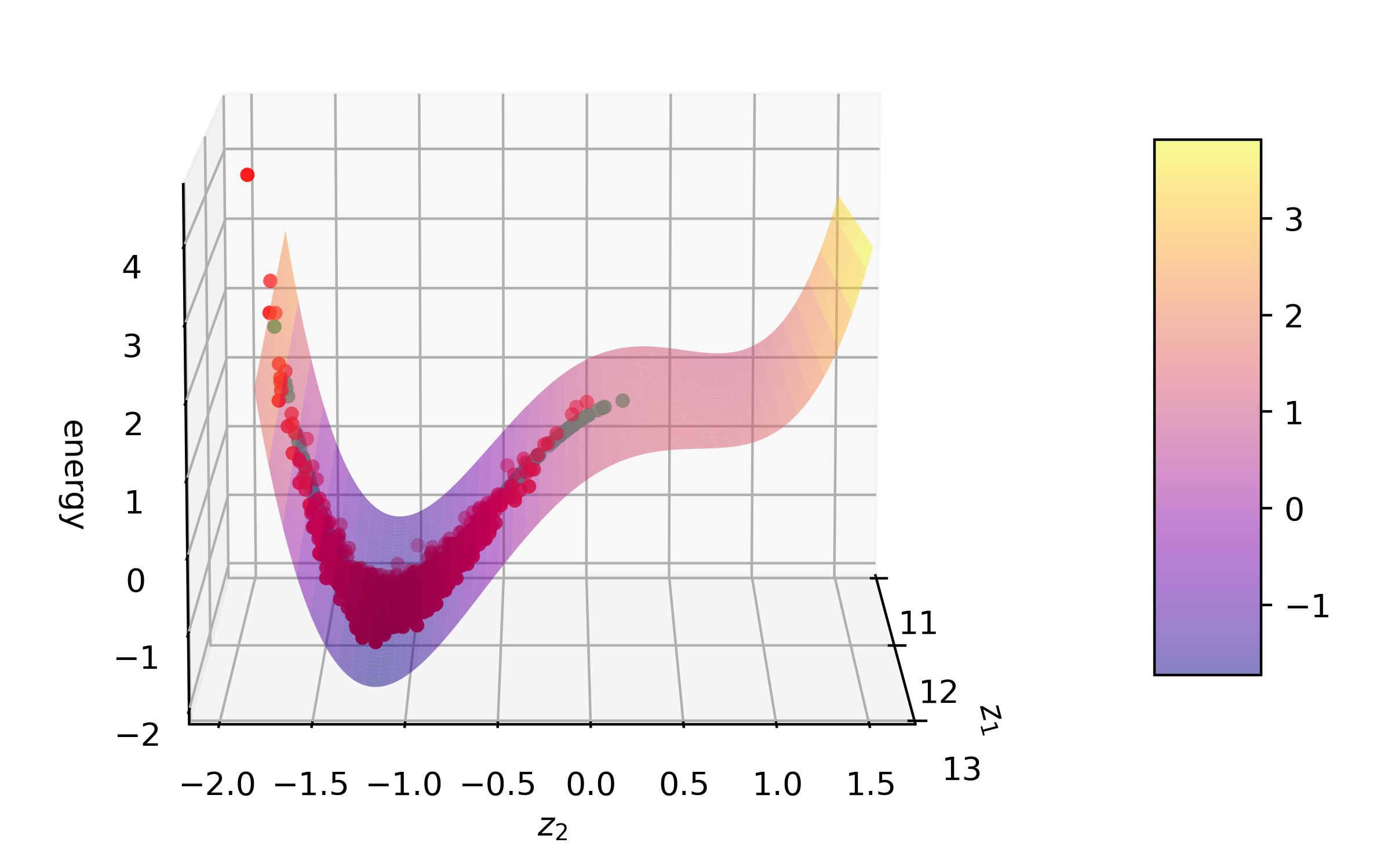}
    \caption{The output from a trained SGM conditioned at $z_1 = 12$ (red) and a US simulation output (green) shown in Figure \ref{fig:SGM_and_US_comp_12} are plotted against the potential energy surface of the system for a range around the conditioned $z_1$ value. Both the vertical axis and color of the surface describe the potential energy of the system. Note that there is now one local minimum at $z_2 = -1$ and the former local minimum at $z_2 = 1$ has flattened. Thus, both the SGM and US simulation mostly produce values around the single local minimum.}
    \label{fig:SGM_and_US_12_FES}
\end{figure}

Figure \ref{fig:SGM_and_US_comp_5} shows that, once trained, SGMs can provide data that is a valid approximation of a prescribed conditional distribution in a dynamical system. Additionally, cSGMs are capable of extrapolating beyond their training data. Figure \ref{fig:SGM_and_US_comp_12} demonstrates that SGMs can step out from their training data and {\em provide a valid extrapolated conditional distribution} of data. Figure \ref{fig:SGM_and_US_12_FES} shows that the extrapolation produces values consistent with the physics and the free energy surface of the system. This is useful for dynamical systems with trajectories that are computationally expensive to generate, or systems with large free energy barriers between relative minima. However, the outputs of the cSGM are ultimately approximations, and do not truly sample the physics of the system in question. Thus, the need for the framework  proposed in Section \ref{sec:alg} arises.

\section{SGMs for Initial Condition Generation}
\label{sec:SGM_ICs}

\subsection{SGM Assisted Physics-based Sampling}
Consider a dynamical system described by the following two-dimensional set of multiscale SDEs:
\begin{equation}
    \begin{split}
        \mbox{d} x_1(t) &= a_1\mbox{d} t + a_2\mbox{d} B_1 \\
        \mbox{d} x_2(t) &= -((1 + x_2(t))^2 (2 (1 + h - k) x_2(t) - 2 h + 3 (0.75 k - 2) x_2(t)^2 \\
            &+ 4 x_2(t)^3) + 2 (1 + x_2(t)) (h - 2 h x_2(t) + (1 + h - k) x_2(t)^2 \\ &+ (0.75 k - 2) x_2(t)^3 + x_2(t)^4))\mbox{d} t + a_3 \mbox{d} B_2(t)
    \end{split}
\end{equation}
where $B_1$ and $B_2$ are standard normal random variables, $a_1 = a_2 = 10^{-4}$, and $a_3 = \epsilon{a_1} = \epsilon{a_2} = 10^{-1}$ where the ratio of time scales is $\epsilon = 10^3$. Similar to the system detailed in Section \ref{sec:SGM_approx_comp}, $x_1$ is the slow variable and $x_2$ is the fast variable ($x_2$ is faster than $x_1$ by a factor of $10^3$). The behavior of $x_2$ is again defined as a double-well potential where two basins of attraction at $x_2 = 1$ and $x_2 = -1$ are separated by a free energy barrier. However, this system differs from the system in Section \ref{sec:SGM_approx_comp} in that the wells in $x_2$ are not evolving with respect to $x_1$ and, in addition,  in the inclusion of the variable $h$ (that defines the height of the barrier) and the variable $k$ (which describes the depth of the well at $x_2 = 1$). While the two directions, $x_1$ and $x_2$ are now independent, this new system allows us to demonstrate cases which the barrier between the two free energy basins is very large and practically insurmountable without enhanced sampling or high temperature. For the figures and examples shown, $h = 8$ and $k = 0$.  We can also add a harmonic biasing potential to the $x_1$ equation of our system for sampling $x_2$ distributions with US, resulting in the following new equations:
\begin{equation}
    \begin{split}
        \mbox{d} x_1(t) &= (a_1 - \kappa(x_1(t) - x^0_1))\mbox{d} t + a_2\mbox{d} B_1 \\
        \mbox{d} x_2(t) &= -((1 + x_2(t))^2 (2 (1 + h - k) x_2(t) - 2 h + 3 (0.75 k - 2) x_2(t)^2 \\
            &+ 4 x_2(t)^3) + 2 (1 + x_2(t)) (h - 2 h x_2(t) + (1 + h - k) x_2(t)^2 \\ &+ (0.75 k - 2) x_2(t)^3 + x_2(t)^4))\mbox{d} t + a_3 \mbox{d} B_2(t)
    \end{split}
    \label{eq:dwell_biased}
\end{equation}
\noindent
where $\kappa$ is a force constant dictating the strength of the bias, and $x^0_1$ is the value of $x_1$ to bias toward. In order to produce initial conditions for the US simulations, we train an SGM on a long unbiased simulation of this system where the samples are labeled by their $x_1$ values (in this case, the dataset consisted of 100,000 samples from a simulation of these SDEs with $\mbox{d} t = 10^{-2}$). Once trained, our SGM can generate point distributions in space consistent with a prescribed $x_1$ value (as demonstrated in Section \ref{sec:SGM_approx_comp}). We then use these generated points to initialize ten US simulations parameterized by \eqref{eq:dwell_biased}. The ten US simulations were run with $\mbox{d} t = 10^{-2}$ for $1000$ time steps each.

\begin{figure}[ht]
    \centering
    \includegraphics[width=1\linewidth]{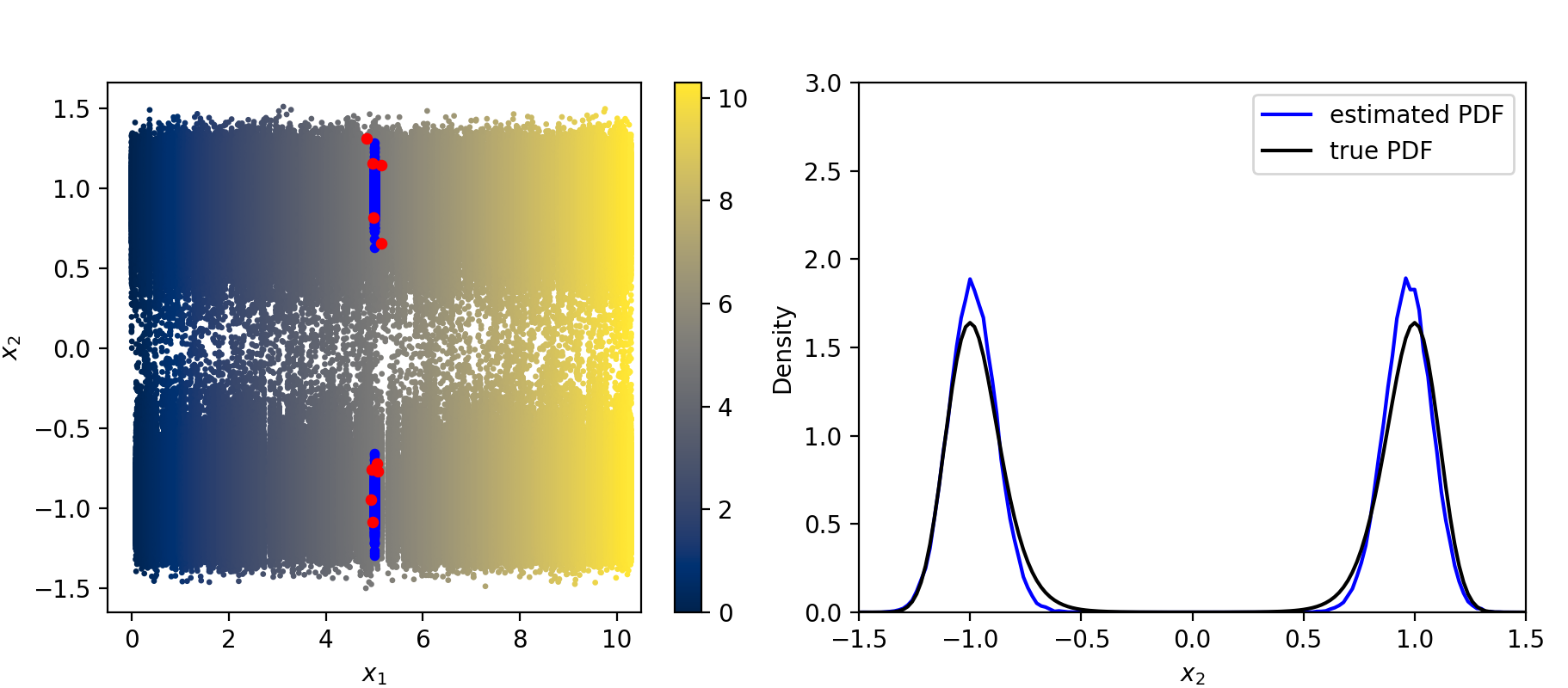}
    \caption{On the left, outputs of 10 US simulations of $1,000$ time steps (blue) conditioned at $x_1 = 5$ initialized by SGM generated points (red) on top of a background of the training data for the SGM. On the right, the probability density function estimated from pooling the histograms of the coupled US+SGM simulations (blue line) compared to the true probability density function given by the known $x_2$ equation of the system (black line).}
    \label{fig:US_outputs}
\end{figure}

\begin{figure}[ht]
    \centering
    \includegraphics[width=0.9\linewidth]{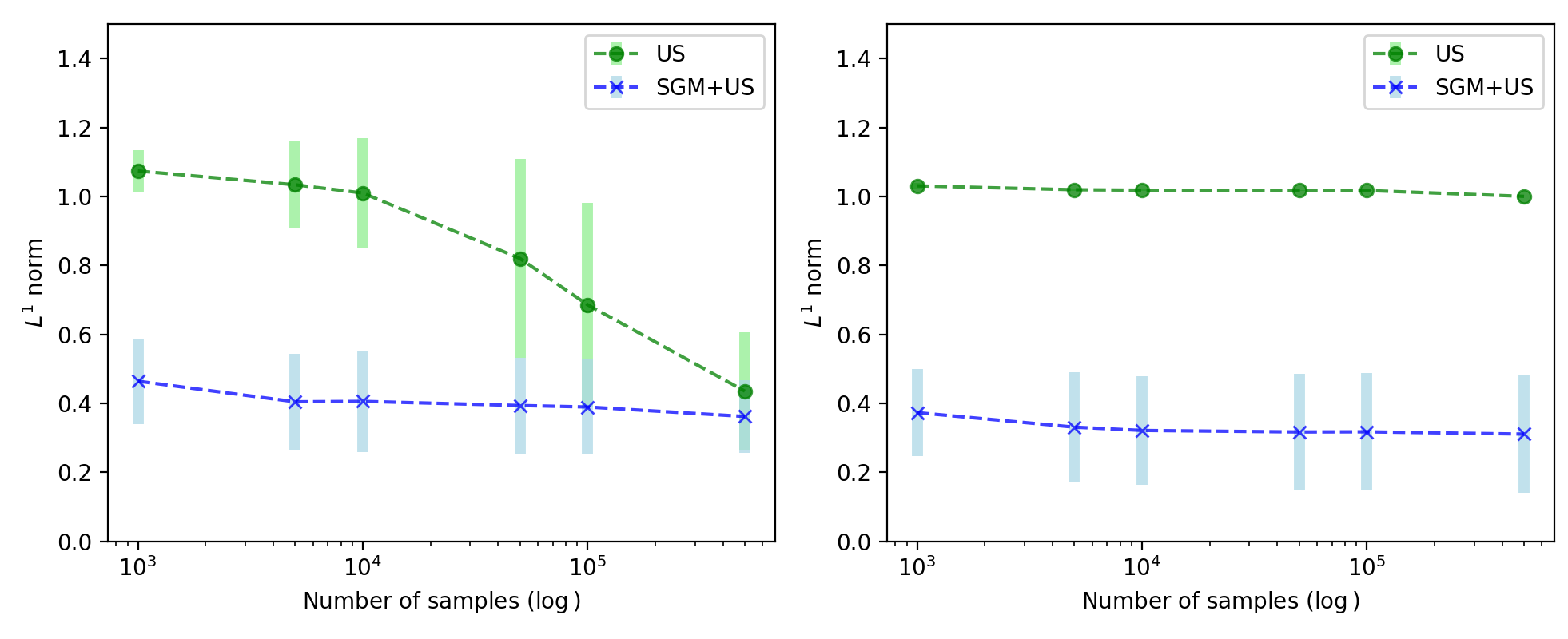}
    \caption{The $L^1$ norm of the absolute value of the difference between the true PDF and the estimated PDF determined by US (green) and coupled SGM+US (blue) runs at two different barrier heights. On the left, the barrier height is $h = 4$. On the right, $h = 8$ , $L^1$ norm is a function of the sample size (\emph{i.e.,} number of time steps). Each data point and its corresponding error bar reflects the results of 1,000 independent experiments.}
    \label{FES}
\end{figure}

These US simulations produce conditional measures of $x_2$ values constrained on $x_1$ values that are close to the respective biased value, $x^0_1$. If the true distribution of the unbiased variable ($x_2$ in this case) is of interest, each US run may not fully span the true conditional distribution, so it is necessary to combine the histograms of each US run in order to produce the true conditional distributions of the system, which is accomplished through pooling the histograms. 
If biases are introduced to the $x_2$ direction, reweighting techniques such as the weighted histogram analysis method (WHAM) \cite{wham} (or multistate Bennett acceptance ratio (MBAR) \cite{mbar}) can be used to combine the biased histograms. These biased histograms are then utilized through WHAM to approximate the free energy surface (and thus, the unbiased joint distribution of $x_2$).

In addition to the multiple runs being able to be run \emph{in parallel}, we can also perform convergence analysis with respect to the number of integration steps of the discretized system of SDEs required for convergence to the true probability density of $x_2$. Figure 2 shows that the coupling of a SGM with US results in a further increased performance, as compared to US alone, especially at smaller sample sizes, because US alone can often become trapped in the metastable states of the system.

\subsection{Manifold Learning For Identifying Collective Variables}
\label{sssec:dmaps_SGM}

Manifold learning or dimensionality reduction techniques can be used if the slow direction is not known \emph{a priori}, as mentioned in Section \ref{sec:alg}.  
In previous work with conditional generative adversarial networks \cite{GANs_closures} we investigated in some detail the training of generative models on data labeled by values of slow directions identified by nonlinear dimensionality reduction;  we also demonstrated performing biased sampling directly on these data-mined variables. 

This same workflow can now be implemented with SGMs. In order to produce initial conditions for the US simulations, we first run a long unbiased simulation of this system. In this case, the dataset consisted of 10,000 samples from a simulation of these SDEs with $\mbox{d} t = 10^{-2}$. This is downsampled from the previous example for memory considerations for the next dimensionality reduction step. We then use nonlinear dimensionality reduction, in this case Diffusion Maps \cite{COIFMANdmaps}, to identify a coordinate that one-to-one with the slow direction. We finally train the SGM on the data set labeled by the leading diffusion map coordinate $\Phi$.

\begin{figure}[ht]
    \centering
    \includegraphics[width=0.9\linewidth]{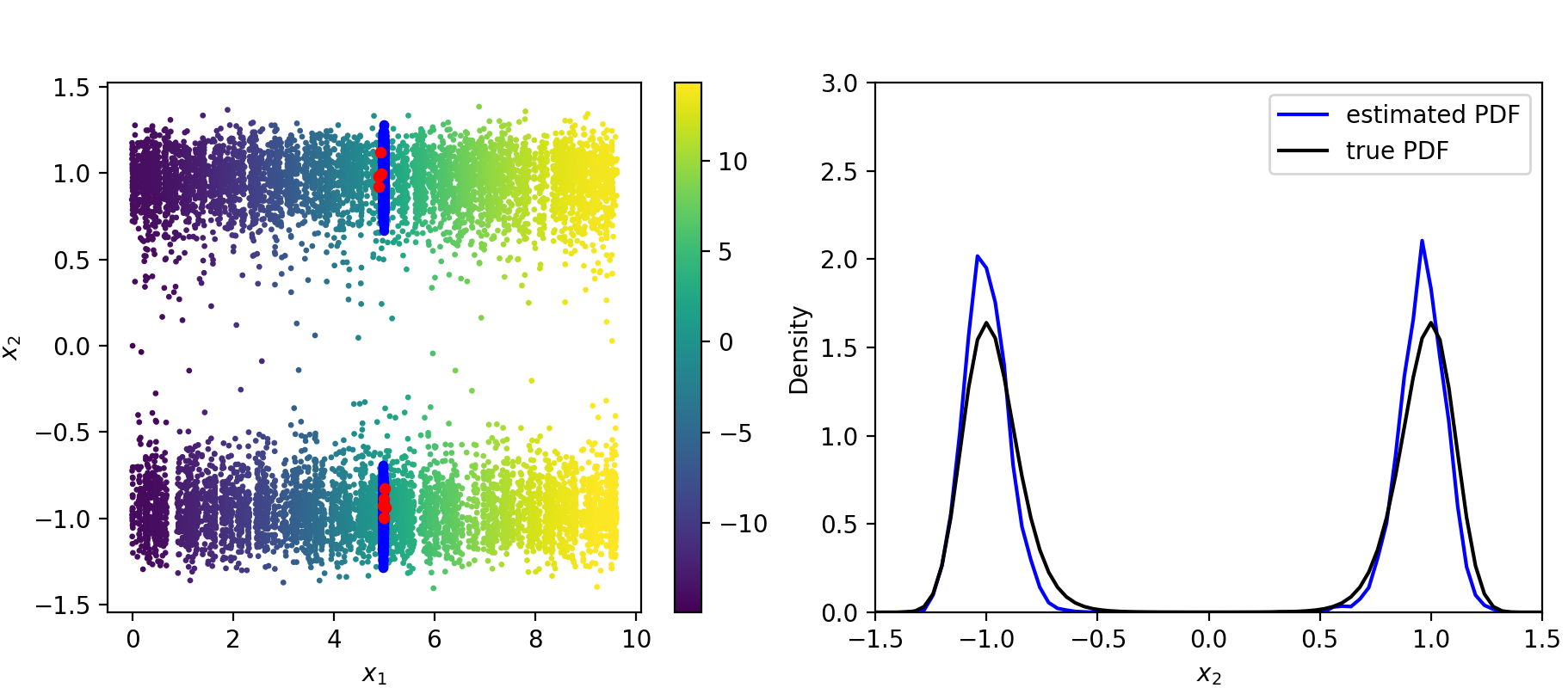}
    \caption{On the left, outputs of 10 US simulations of $1,000$ time steps (blue) conditioned on the Diffusion Map coordinate value $\Phi = 0$ (which corresponds to the approximate halfway point of label produced by diffusion maps that is one-to-one with the slow direction $x_1$) initialized by SGM generated points (red) on top of a background of the training data for the SGM colored by the diffusion map coordinate. The training data is much more sparse than previous examples due to the downsampling performed for the diffusion maps. On the right, the probability density function estimated from pooling the histograms of the coupled US+SGM simulations (blue line) compared to the true probability density function given by the known $x_2$ equation of the system (black line). }
    \label{fig:US_outputs_dmaps}
\end{figure}
The approximated PDF of this example is less accurate than the example labeled by the known slow direction; this should be attributed to the downsampling of the data set and the resulting smaller number of points in the low probability/high energy areas of $x_2$ space.

\subsection{SGM-Generated Initial Conditions for Molecular Dynamics}
SGMs can be trained to generate initial conditions of high dimensional problems. For example in the field of computational chemistry, exploring the conformation space of new molecules of interest is an important step in workflows for drug discovery and materials science. Consider a 20 nanosecond molecular dynamics (MD) simulation of alanine dipeptide performed using the OpenMM molecular dynamics package with the Amber14 force field in implicit solvent, stochastic (Langevin) dynamics performed at 300 Kelvin, a time step of 1 fs, and a collision rate of 1 ps$^{-1}$. \cite{openmm, amber14, GBsolvent} This simulation sampled two local free energy basins at distinct values of $\Phi$ and $\Psi$ (the commonly accepted CVs, which are dihedral angles that characterize the bulk behavior of the molecule). We then train an SGM on the molecular conformations from this simulation labeled by their values of $\Psi$. The result is an SGM that can generate alanine dipeptide molecules that are consistent with a desired CV value. These in turn can be used as proposed in section 4.1 to initialize US simulations and approximate free energy surfaces along the CV coordinate through the use of WHAM. Note that SGMs have been used to generate molecular conformations \cite{diffdock, molSGM}, but coupling SGMs with physics-based models has not yet been investigated, to our knowledge. All parameters of the SGM SDE were kept consistent with previous examples, as were the learning rates; however, the score network that was used for these molecules was larger than the previous networks for the SDE dynamical system examples.

\begin{figure}[H]
    \centering
    \includegraphics[width=0.9\linewidth]{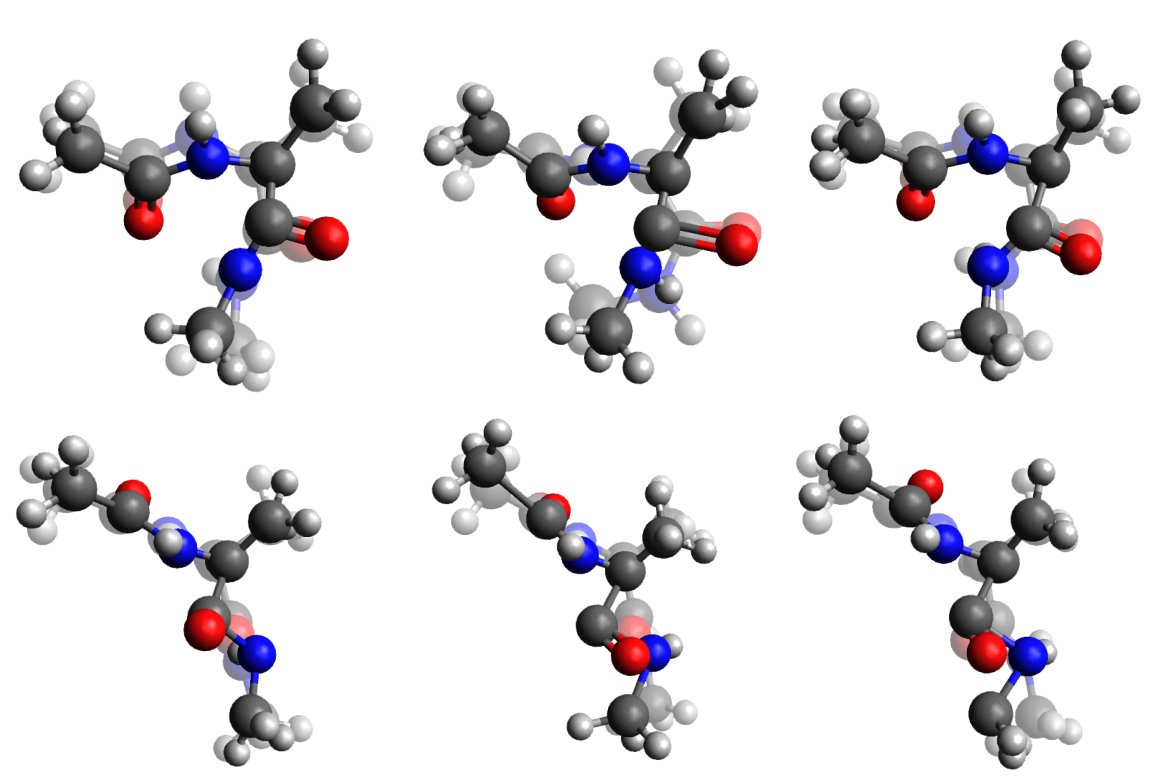}
    \caption{Alanine dipeptide molecular conformations produced by an SGM. In the top row are alanine molecules which have a $\Psi$ value equal to approximately 4 radians. In the bottom row, the alanine molecules $\Psi$ value is equal to approximately 6 radians. The transparent conformation visible ``behind" each molecule is the same molecule after its energy is minimized through 50 steps of a steepest descent algorithm.}
    \label{fig:alanine}
\end{figure}

\begin{figure}[H]
    \centering
    \includegraphics[width=0.75\linewidth]{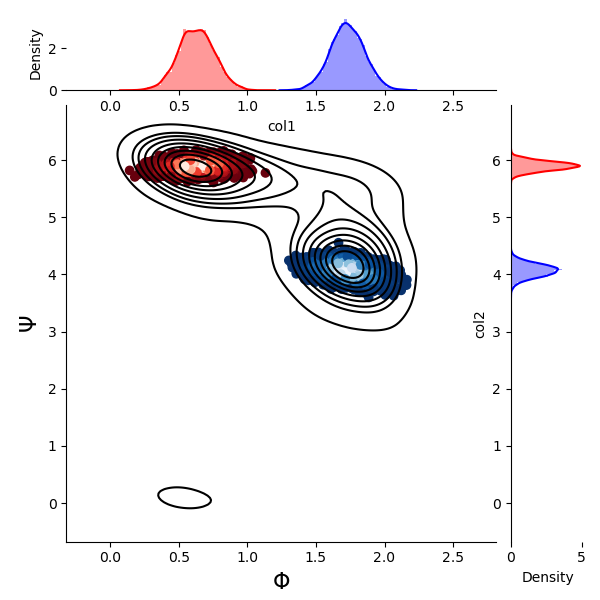}
    \caption{Dihedral angles of many SGM output samples; red(blue) correspond to conformations produced by a SGM conditioned on the $\Psi$ value of the left(right) free energy minimum. Color gradients (dark red to light red, dark blue to light blue) represent empirical sample densities. The black contour lines represent the empirical sample density of a long unbiased simulation run that was used to train the SGM. The empirical marginals for the SGM samples are included on the margins.}
    \label{fig:ala46}
\end{figure}

\section{Conclusions}
\label{sec:conclusions}
Conditional SGMs, like other conditional generative models, exhibit considerable parallels with physics-based sampling methods. Coupling the ML-based SGMs with established approaches in physics and computational chemistry can mitigate the need for guessing appropriate initial conditions for simulation and can also alleviate the need for long equilibration simulations starting at such initial conditions. Moreover, coupling SGMs with US (and WHAM where relevant) provides a faster convergence to the true joint distribution when compared to US alone (and may even provide convergence to the joint distribution when US alone will not realistically converge). In future work, we hope to use this framework to investigate systems with more complex equilibrium distributions. In particular, we hope to investigate more complex molecular model-based multiscale systems as well as dynamical systems in which bifurcations are present.

\begin{acknowledgments}
The authors wish to acknowledge the support of the US Department of Energy and the Air Force Office of Scientific Research (through an AFOSR MURI). We also thank Holden Lee for insightful discussions.
\end{acknowledgments}

\section*{Data Availability Statement}
The data and code that supports the findings of this study are
openly available in GitHub at \url{https://github.com/ecrab/SGM_assisted_sampling}

\appendix
\section{SGMs with Manifold Learning Continued}
 In subsection \ref{sssec:dmaps_SGM}, we showed that, if the CVs of a dynamical system of interest are not known \emph{a priori}, one could use nonlinear dimensionality reduction to identify low dimension variables that are one-to-one with the CVs of the system. Consider again the example used in \ref{sec:SGM_approx_comp}. Instead of using the slow direction, $z_1$, as the label for the cSGM, we use diffusion maps to identify a leading eigenvector coordinate $\Phi_1$ that is one-to-one with $z_1$. The training data used for this model is more sparse (by a factor of 20) than the original example in \ref{sec:SGM_approx_comp} due to the cost of calculating the transition matrix for diffusion maps.

 \begin{figure}[H]
    \centering
    \includegraphics[width=1\linewidth]{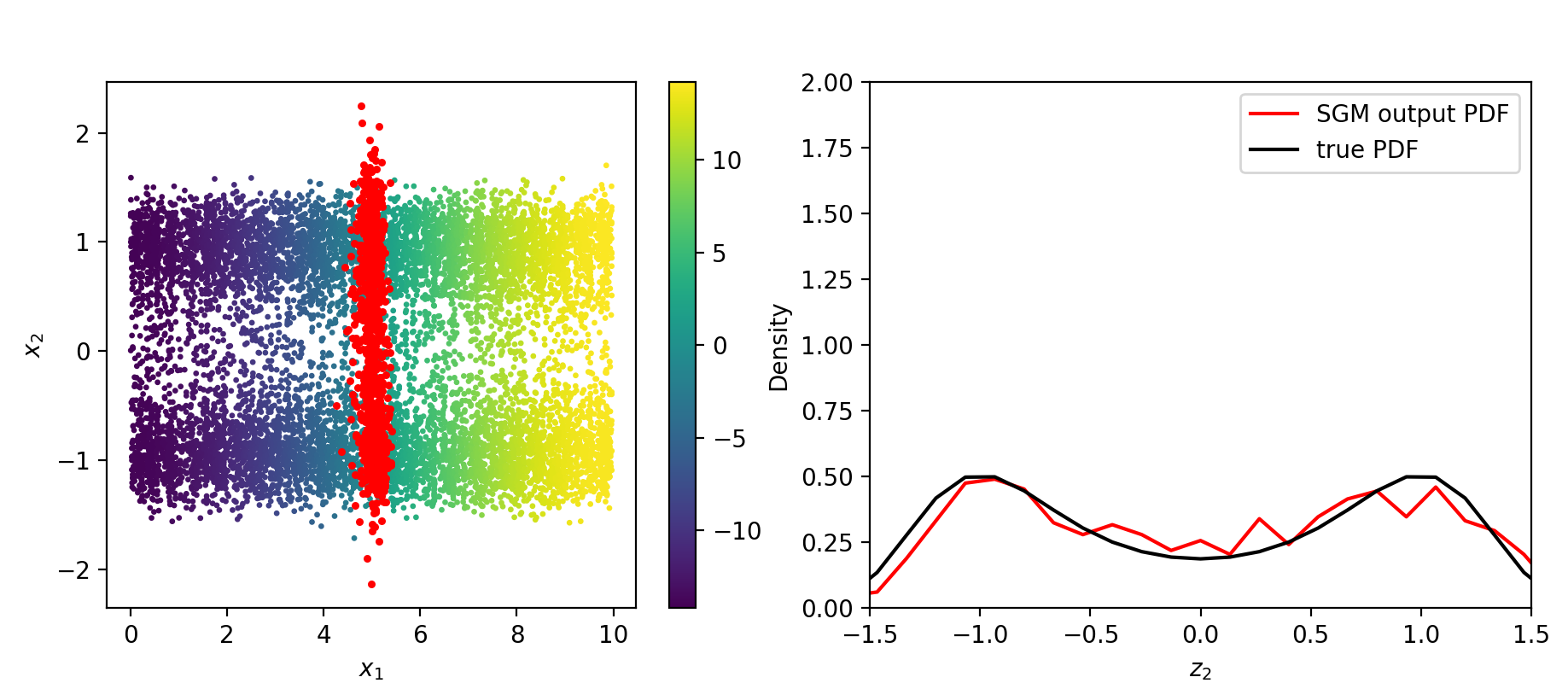}
    \caption{On the left, output from a trained SGM conditioned on the leading Diffusion Maps coordinate value $\Phi_1 = 0 $ (red) on top of a background of the training data for the SGM colored by $\Phi_1$ values. On the right, the PDF estimated from the SGM output is compared to the true PDF given by the known $z_2$ equation of the system (black line).}
    \label{fig:dmaps_comp}
\end{figure}

\bibliographystyle{unsrt}
\bibliography{bibliography}
\end{document}